 \documentclass[pmlr,twocolumn,10pt]{jmlr} 

\usepackage{booktabs}
\usepackage{siunitx}

\usepackage[switch]{lineno}

\theorembodyfont{\upshape}
\theoremheaderfont{\scshape}
\theorempostheader{:}
\theoremsep{\newline}

\jmlrvolume{LEAVE UNSET}
\jmlryear{2024}
\jmlrsubmitted{LEAVE UNSET}
\jmlrpublished{LEAVE UNSET}
\jmlrworkshop{Machine Learning for Health (ML4H) 2024} 

\title[A Demonstration of Adaptive Collaboration of Large Language Models for Medical Decision-Making]{A Demonstration of Adaptive Collaboration of Large Language Models for Medical Decision-Making}

\author{%
    \Name{Yubin Kim}$^1$ \Email{ybkim95@mit.edu} \\
    \Name{Chanwoo Park}$^1$ \Email{cpark97@mit.edu} \\
    \Name{Hyewon Jeong}$^1$ \Email{hyewonj@mit.edu} \\
    \Name{Cristina Grau-Vilchez}$^1$ \Email{crisgrau@media.mit.edu} \\
    \Name{Yik Siu Chan}$^1$ \Email{yiksiuc@mit.edu} \\
    \Name{Xuhai Xu}$^2$ \Email{xx2489@columbia.edu} \\
    \Name{Daniel McDuff}$^3$ \Email{dmcduff@google.com} \\
    \Name{Hyeonhoon Lee}$^4$ \Email{hhoon@snu.ac.kr} \\
    \Name{Cynthia Breazeal}$^1$ \Email{cynthiab@media.mit.edu} \\
    \Name{Hae Won Park}$^1$ \Email{haewon@media.mit.edu} \\
    \\
    \addr $^1$ Massachusetts Institute of Technology \\
    \addr $^2$ Columbia University \\
    \addr $^3$ Google Research \\
    \addr $^4$ Seoul National University Hospital \\
}

\begin{document}
\vspace{-15pt}

\maketitle

\vspace{-10pt}

\section{Introduction}

\vspace{-10pt}

Medical Decision-Making (MDM) is a multi-faceted process that requires clinicians to assess complex multi-modal patient data patient, often collaboratively~\citep{zhou2023difficulty}. Large Language Models (LLMs)~\citep{openai2024gpt4, Anthropic2024a, reid2024gemini} promise to streamline this process by synthesizing vast medical knowledge and multi-modal health data~\citep{singhal2022largelanguagemodelsencode, kim2024healthllm}. However, single-agent are often ill-suited for nuanced medical contexts requiring adaptable, collaborative problem-solving. Our \textbf{MDAgents}\footnote{Our code and demo can be found at \url{https://mdagents2024.github.io/}}~\citep{kim2024mdagentsadaptivecollaborationllms} addresses this need by dynamically assigning collaboration structures to LLMs based on task complexity, mimicking real-world clinical collaboration and decision-making \cite{parekh2011managing, grembowski2014conceptual}. This framework improves diagnostic accuracy and supports adaptive responses in complex, real-world medical scenarios, making it a valuable tool for clinicians in various healthcare settings, and at the same time, being more efficient in terms of computing cost  than static multi-agent decision making methods.

\begin{figure*}[t!]
  \centering
  \includegraphics[width=1.0\textwidth]{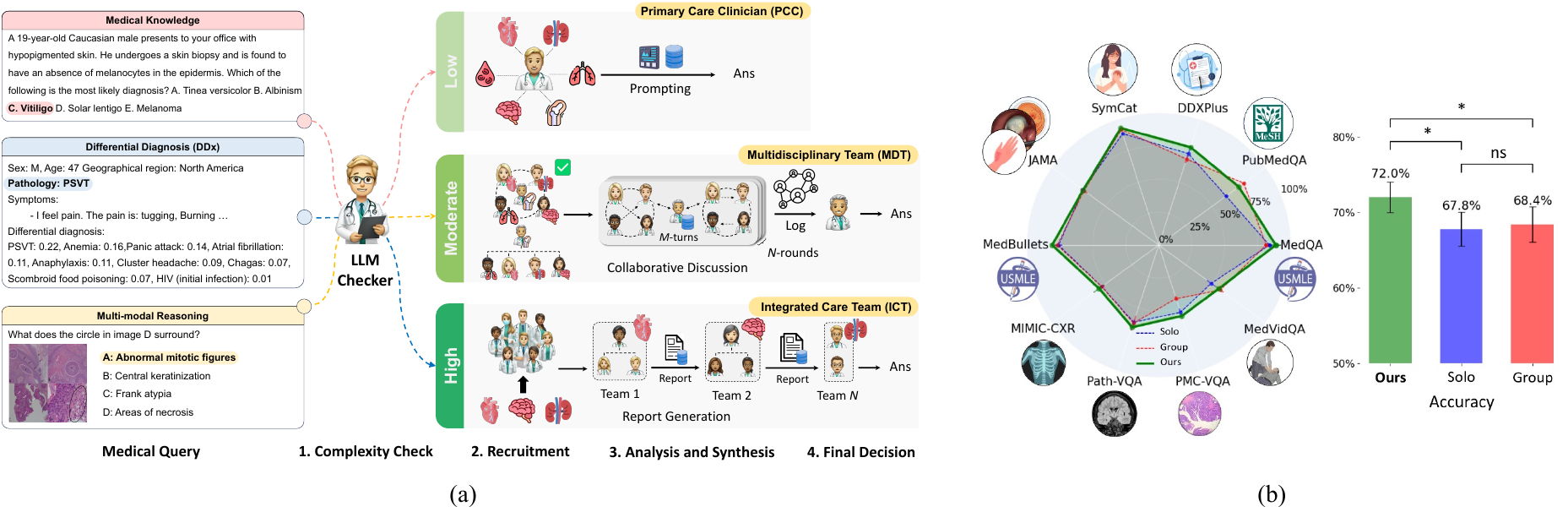}
  \caption{\textbf{Overview.} \textbf{(a)} The system assesses complexity, recruits agents, and iteratively refines answers through solo, MDT, and ICT configurations, emulating real-world clinical collaborations. \textbf{(b)} Our method outperforms Solo and Group settings across different medical benchmarks.}
  \label{fig:framework}
\end{figure*}

\section{Method}
As shown in Figure \ref{fig:framework}-(a), MDAgents operates in four steps, adapting to each medical query:

\begin{enumerate}
    \item \textbf{Complexity Check}: A \textit{moderator agent} evaluates the query and assigns a complexity level (\textit{low}, \textit{moderate}, or \textit{high}) based on established clinical guidelines.
    \vspace{-3pt}
    \item \textbf{Recruitment}: A \textit{recruiter agent} then forms an appropriate team. Low-complexity cases go to a single Primary Care Physician (PCP), while more complex cases have Multi-disciplinary Teams (MDTs) or Integrated Care Teams (ICTs).
    \vspace{-3pt}
    \item \textbf{Analysis and Synthesis}: Solo agents use techniques like Chain-of-Thought (CoT)~\citep{wei2022chain} prompting for analysis, while MDTs refine answers through rounds of discussion. ICTs use a tiered decision process for the high complexity cases, engaging specialists in sequential stages.
    \vspace{-3pt}
    \item \textbf{Final Decision}: The decision-maker agent aggregates inputs and synthesizes a final answer, leveraging diverse agent insights, conversational history, and moderator feedback.
\end{enumerate}
For up-to-date decision-making, MDAgents incorporates \textbf{MedRAG}~\citep{xiong2024benchmarking}, which accesses recent biomedical data to enhance accuracy. The model leverages GPT-4 for reasoning and was tested across medical benchmarks, including text-only (e.g., MedQA~\citep{jin2020medqa}), image-based (e.g. Path-VQA~\citep{he2020pathvqa}), and multi-modal (e.g. MedVidQA~\citep{gupta2022dataset}) datasets.

To evaluate the effectiveness of our framework, we conducted extensive experiments using baseline methods across ten datasets, including MedQA, PubMedQA, and others. Each dataset was tested with 50 samples, and we measured the average inference times for varying complexities: low (14.7 seconds), moderate (95.5 seconds), and high (226 seconds). Our experiments compared three configurations: (1) Solo, utilizing a single LLM agent for decision-making; (2) Group, where multiple agents collaborate; and (3) Adaptive, our proposed MDAgents method, which dynamically adjusts the inference structure. For low-complexity cases, we employed 3-shot prompting, while moderate and high-complexity cases used zero-shot prompting across all settings. 

\vspace{-15pt}

\section{Results}

\vspace{-5pt}

As shown in Figure \ref{fig:framework}-(b), MDAgents demonstrated state-of-the-art performance, achieving the highest accuracy in \textbf{7 out of 10} benchmarks and outperforming both solo LLMs and static multi-agent methods. Across different medical benchmarks, MDAgents achieved up to 4.2\% improvement over solo and group setting. From our ablations in Figure \ref{fig:ablation2}-\ref{fig:consensus} in Appendix, we present MDAgent requires fewer API calls than larger static multi-agent models, achieving the same accuracy with fewer resources. Specifically, using 3 agents yielded optimal performance with a significantly lower number of API calls than single-agent CoT or larger multi-agent setups (e.g., MedAgents with 5 agents), highlighting our framework's computational efficiency. We also present that our framework shows robust performance in extreme temperature values (\textit{T}=0.3, \textit{T}=1.2) (Figure \ref{fig:ablation2}) and the agent's opinion convergence in multi-agent collaboration settings (Figure \ref{fig:consensus}).

\vspace{-15pt}

\section{Discussion}

\vspace{-5pt}

Developing MDAgents required balancing agent count and response accuracy while minimizing resource use. The framework’s effectiveness lies in its adaptive decision-making, which \textbf{mirrors real clinical practice} by scaling collaboration with case complexity. Ablations showed that a 3-agent setup optimizes performance, minimizing unnecessary complexity for simpler cases while supporting complex ones. Key insights include verification steps to prevent hallucinations and the use of tools like MedRAG and moderator oversight to enhance diagnostic accuracy. MDAgents is currently being evaluated for integration in hospital settings to support clinical workflows.

\paragraph{Future Directions.} We plan to integrate doctor-in-the-loop feedback, keeping MDAgents aligned with clinical knowledge and enhancing reliability to reduce diagnostic errors and improve patient outcomes.

\acks{C.P. acknowledges support from the Takeda Fellowship, the Korea Foundation for Advanced Studies, and the Siebel Scholarship.}

\bibliography{jmlr-sample}

\appendix


\begin{figure*}[h!]
  \centering
  \includegraphics[width=1.0\textwidth]{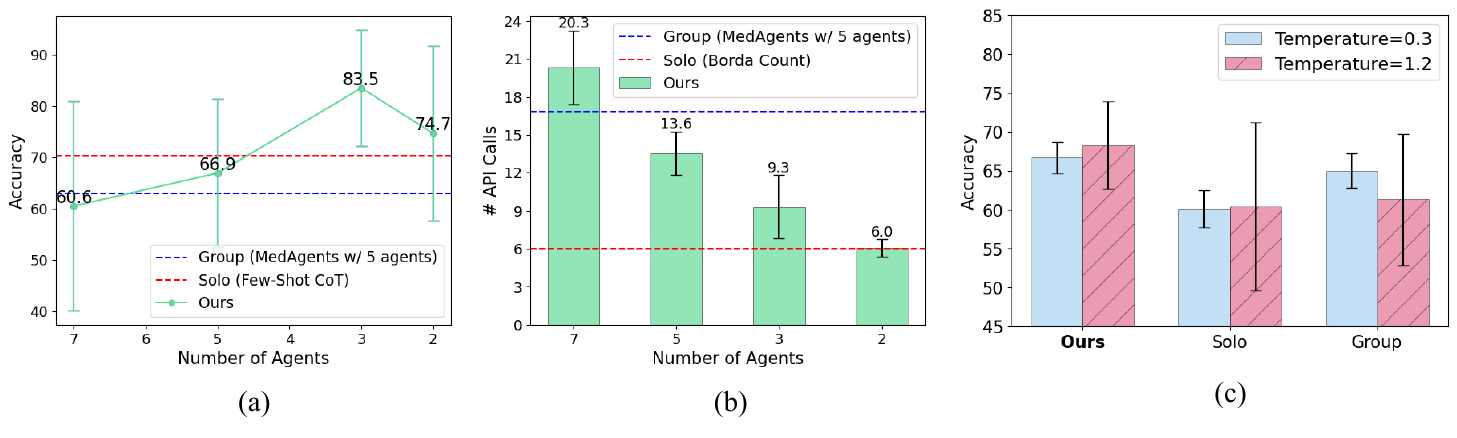}
  \caption{Impact of the number of agents on (a) Accuracy, (b) Number of API Calls on medical benchmarks with GPT-4 (V) and (c) Performance of three different settings under low (\textit{T}=0.3) and high (\textit{T}=1.2) temperatures on medical benchmarks. Our Adaptive setting shows better robustness to different temperatures and even takes advantage of higher temperatures.}
  \label{fig:ablation2}
\end{figure*}

\begin{figure*}[t!]
  \centering
  \includegraphics[width=0.6\textwidth]{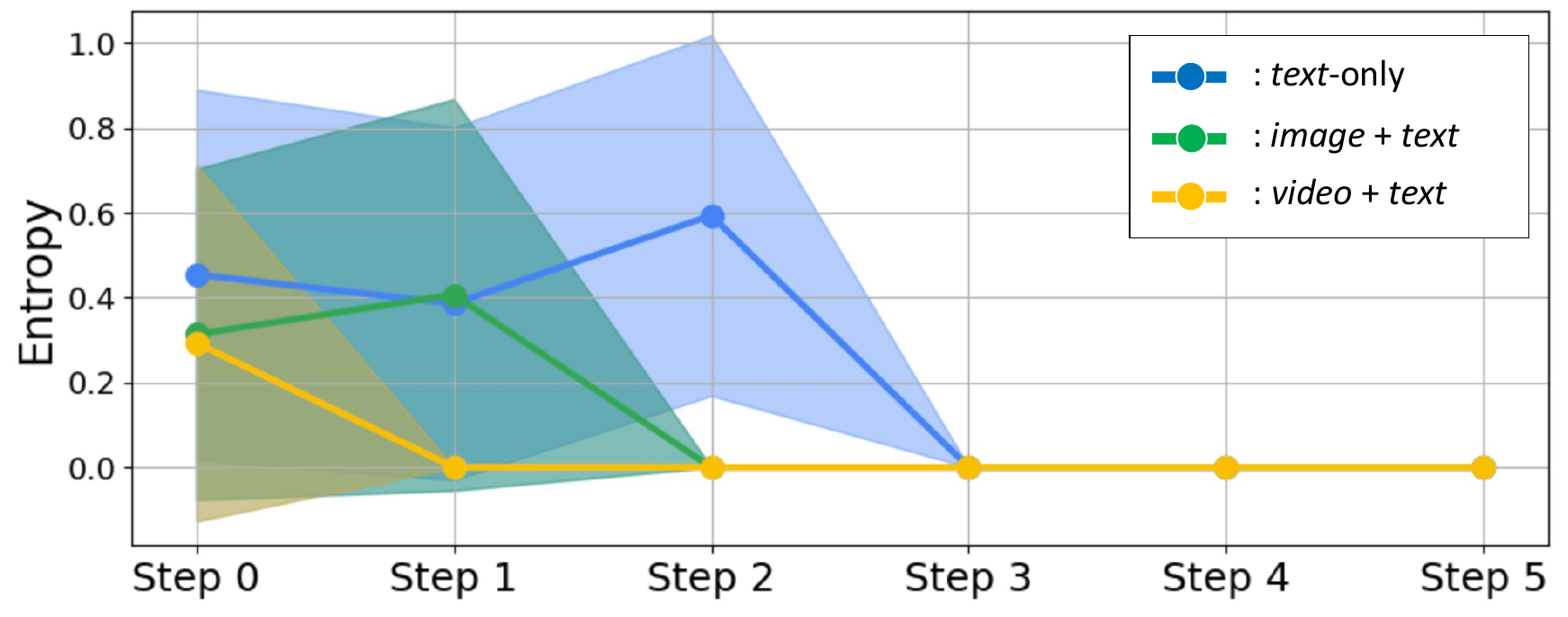}
  \caption{An illustration of consensus entropy in group collaboration process of MDAgents (w/ Gemini-Pro (Vision), \textit{N}=30 for each dataset) on medical benchmarks with different modality inputs.}
  \label{fig:consensus}
\end{figure*}

\section{Demonstration Details}\label{apd:demo}

\begin{figure*}[t!]
  \centering
  \includegraphics[width=1.0\textwidth]{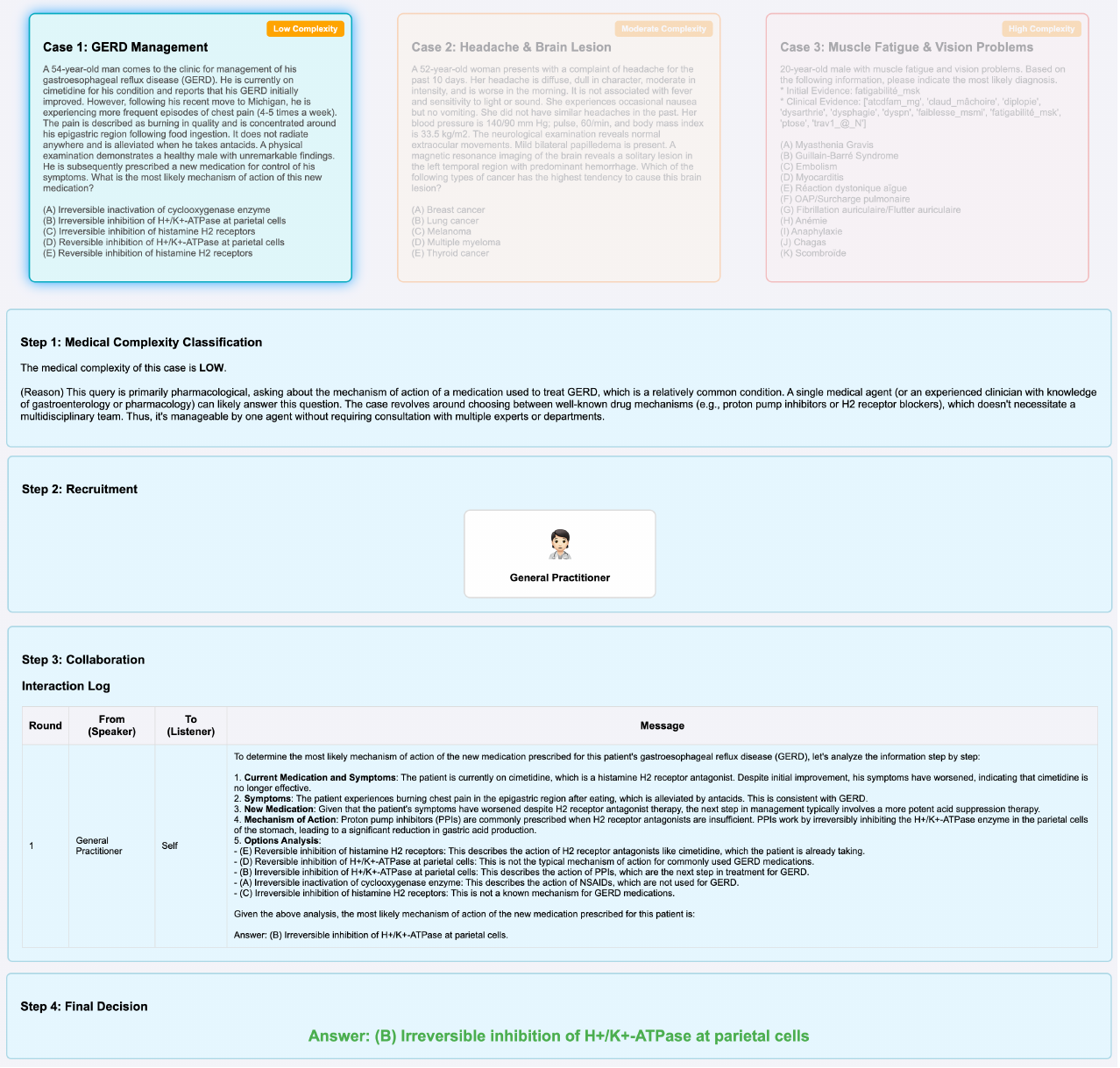}
  \caption{In a \textit{\textbf{low}}-complexity case (GERD management), the MDAgents framework assigns a single general practitioner agent to handle a straightforward pharmacological query independently.}
  \label{fig:demo2}
\end{figure*}

\begin{figure*}[t!]
  \centering
  \includegraphics[width=1.0\textwidth]{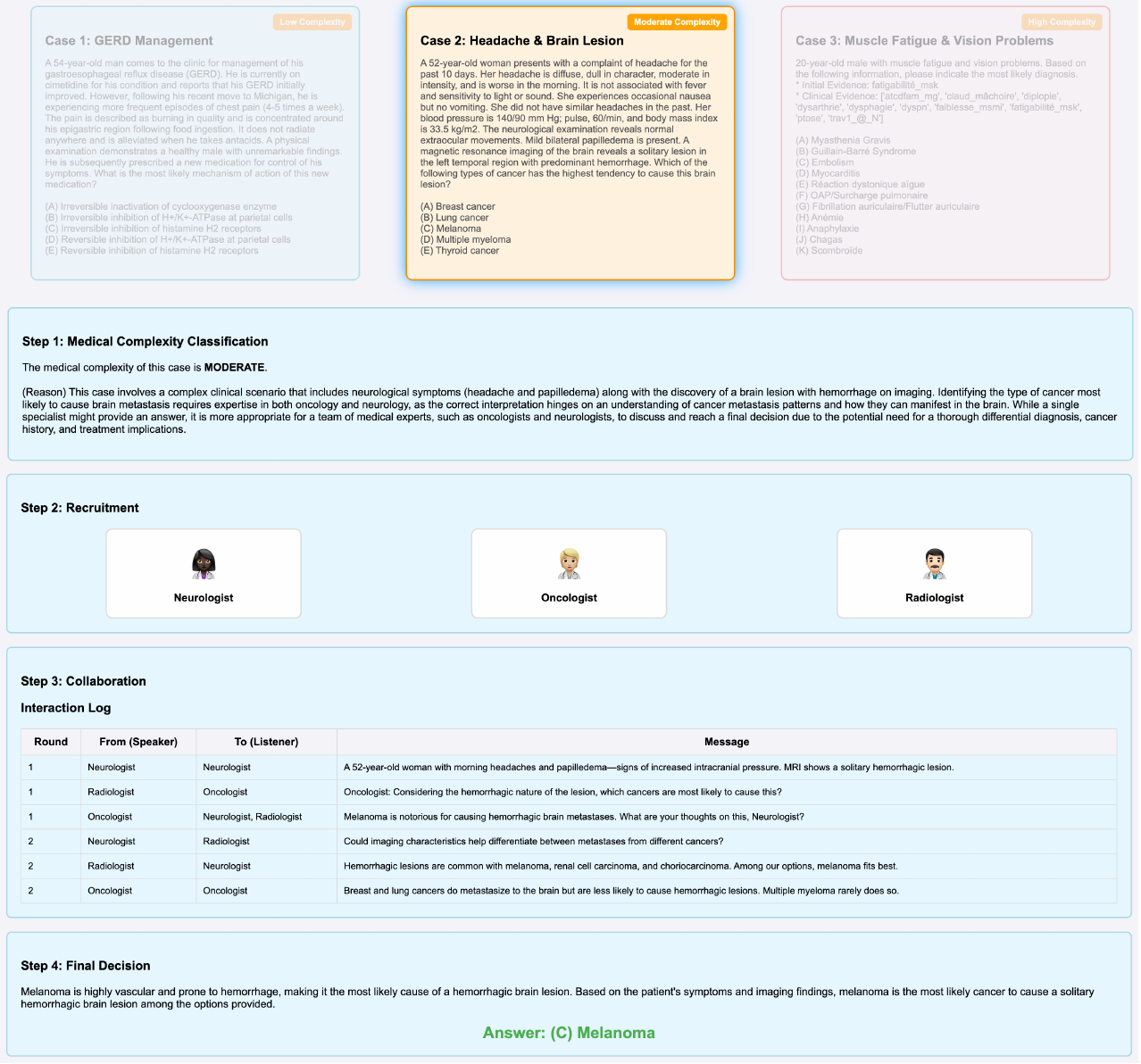}
  \caption{In a \textit{\textbf{moderate}}-complexity case (Headache \& Brain Lesion), the framework recruits multiple agents (MDT)—a neurologist, oncologist, and radiologist—for collaborative reasoning to diagnose the likely cause of a brain lesion, requiring specialized input.}
  \label{fig:demo3}
\end{figure*}

\begin{figure*}[t!]
  \centering
  \includegraphics[width=1.0\textwidth]{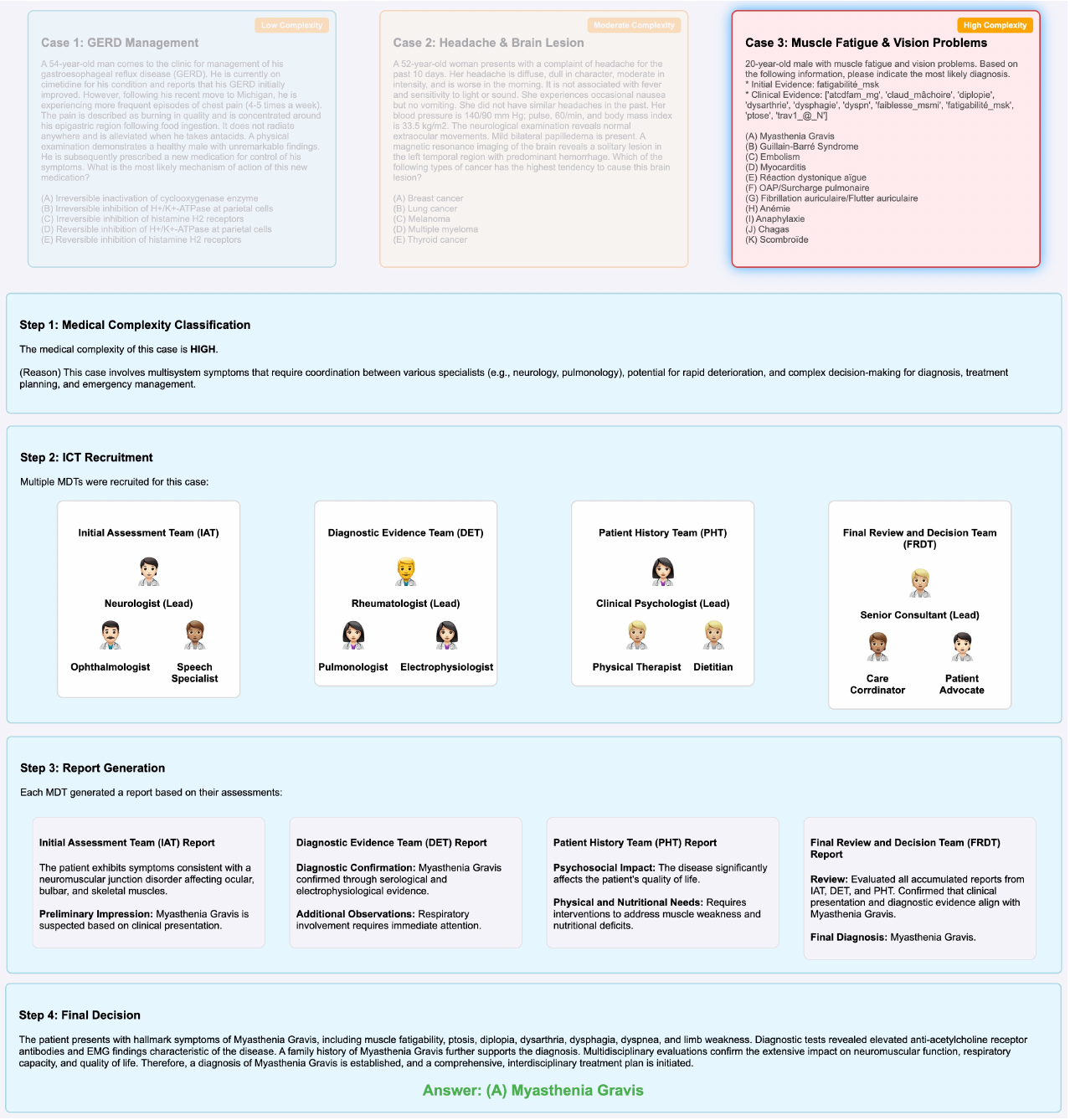}
  \caption{In a \textit{\textbf{high}}-complexity case (Muscle Fatigue \& Vision Problems), MDAgents coordinates multiple MDTs (i.e. ICT) across disciplines (e.g., neurology, pulmonology, psychology) to address complex, multisystem symptoms, reflecting a comprehensive interdisciplinary approach.}
  \label{fig:demo4}
\end{figure*}

The demonstration, accessible at \url{https://mdagents2024.github.io/demo.html}, showcases the MDAgents framework's capability to adapt to cases of varying complexity. For each case:
\begin{itemize}
    \item \textbf{Low Complexity (Figure \ref{fig:demo2}):} A single general practitioner agent manages a straightforward query on GERD management.
    \item \textbf{Moderate Complexity (Figure \ref{fig:demo3}):} The framework recruits multiple agents (MDT)—a neurologist, oncologist, and radiologist—for collaborative reasoning in a case involving headache and brain lesion.
    \item \textbf{High Complexity (Figure \ref{fig:demo4}):} MDAgents coordinates several specialized teams (ICT) across disciplines (e.g., neurology, pulmonology, psychology) to tackle a multisystem case of muscle fatigue and vision problems, emphasizing interdisciplinary collaboration.
\end{itemize}

Currently, this demo is powered by pre-collected LLM responses, illustrating how MDAgents works in different scenarios. For users interested in real-time usage, demos can be executed by following the instructions at \url{https://github.com/mitmedialab/MDAgents}.

\end{document}